\documentclass{llncs}
\usepackage[utf8]{inputenc}
\usepackage[nolist]{acronym}
\usepackage{amsfonts,epsfig,latexsym,amssymb,yhmath,hyperref}
\usepackage{booktabs}
\usepackage{array}
\usepackage{makeidx}  
\usepackage[usenames,dvipsnames]{color}
\usepackage{cleveref}
\usepackage{float}

\newcolumntype{L}[1]{>{\raggedright\let\newline\\\arraybackslash\hspace{0pt}}m{#1}}
\newcolumntype{C}[1]{>{\centering\let\newline\\\arraybackslash\hspace{0pt}}m{#1}}
\newcolumntype{R}[1]{>{\raggedleft\let\newline\\\arraybackslash\hspace{0pt}}m{#1}}

\title{
Fully Automated Segmentation of Hyperreflective Foci in Optical Coherence Tomography Images
}

\author{Thomas Schlegl\inst{1,2}, Hrvoje Bogunovic\inst{2} \and Sophie Klimscha\inst{2} \and Philipp Seeb\"ock\inst{1,2} \and Amir Sadeghipour\inst{2} \and Bianca Gerendas\inst{2} \and Sebastian M. Waldstein\inst{2} \and \\Georg Langs\inst{1}\thanks{Corresponding author: Georg Langs, Department of Biomedical Imaging and Image-guided Therapy, Computational Imaging Research Lab, Medical University of Vienna, Waehringer Guertel 18-20, 1090, Wien, Austria. georg.langs@meduniwien.ac.at; Thomas Schlegl: thomas.schlegl@gmail.com} \and Ursula~Schmidt-Erfurth\inst{2}
}
\institute{$^1$Department of Biomedical Imaging and Image-guided Therapy, Computational Imaging Research Lab, Medical University Vienna, Austria\\
$^2$Christian Doppler Laboratory for Ophthalmic Image Analysis, Department of Ophthalmology and Optometry, Medical University Vienna, Austria}

\begin{document}

\maketitle

\begin{abstract}
The automatic detection of disease related entities in retinal imaging data is relevant for disease- and treatment monitoring. It enables the quantitative assessment of large amounts of data and the corresponding study of disease characteristics. The presence of hyperreflective foci (HRF) is related to disease progression in various retinal diseases. Manual identification of HRF in spectral-domain optical coherence tomography (SD-OCT) scans is error-prone and tedious. We present a fully automated machine learning approach for segmenting HRF in SD-OCT scans. Evaluation on annotated OCT images of the retina demonstrates that a residual U-Net allows to segment HRF with high accuracy.
As our dataset comprised data from different retinal diseases including age-related macular degeneration, diabetic macular edema and retinal vein occlusion, the algorithm can safely be applied in all of them though different pathophysiological origins are known.

\end{abstract}

\section{Introduction}
Small, well-defined, dot-shaped lesions with equal or higher reflectivity than the retinal pigment epithelium (RPE), visualized by optical coherence tomography (OCT), have been termed hyperreflective foci (HRF). They have been shown to occur in various retinal diseases, including neovascular age-related macular degeneration, diabetic retinopathy and retinal vein occlusion~\cite{christenbury2013progression,de2015optical,chatziralli2016hyperreflective}. Histopathologic analyses have proposed several etiologies, of which the concept of activated, migrating RPE cells currently seems to be the most widely accepted in the setting of age-related macular degeration~\cite{curcio2017activated}. HRF may also represent lipid exudates, which are frequently seen in diabetic maculopathy~\cite{bolz2009optical}. Numerous studies have linked the presence of HRF to be related to progression of disease~\cite{christenbury2013progression,sleiman2017optical}. Furthermore, location and presence of HRF have been proposed as a negative prognostic factor for visual function~\cite{chatziralli2016hyperreflective,segal2016prognostic}. Given that formation and particularly migration of HRF have been shown to occur in the course of various retinal diseases, robust identification presents the first important step in further exploring the role of these lesions in pathomorphologic disease dynamics. Manual identification and counting of HRF throughout OCT volumes, usually consisting of a few dozen to hundreds of B-scan sections, is error-prone and tedious. Automated identification allows easy analysis of HRF presence, load and dynamics and promotes reproducible research on this highly relevant biomarker.

Here, we apply supervised machine learning and deep neural networks for accurate fully automated HRF segmentation.

\paragraph{Related Work}
Residual networks (ResNet), introduced by He et al.~\cite{he2016deep}, improved the state-of-the-art of diverse visual recognition tasks. They won the classification, detection and localization challenge of the in the \textit{ImageNet Large Scale Visual Recognition Challenge (ILSVRC)}, and the \textit{Microsoft Common Objects in Context (MSCOCO)} detection and segmentation challenge. Residual networks ease the training of deep neural networks by introducing residual units, which allow even very deep networks learning the identity mapping.
Deep learning based semantic segmentation approaches, as for example \textit{FCN} of Long et al.~\cite{long2015fully} or \textit{DeepLab} of Chen et al.~\cite{chen2015semantic} have shown to gain higher segmentation accuracy as opposed to the formulation as an image-level convolutional neural network (CNN) based classification problem~\cite{zheng2015conditional}.
Since its introduction by Ronneberger et al.~\cite{ronneberger2015u}, U-Net is a widely used architecture for semantic segmentation. Ronneberger et al. showed that such a network architecture can be trained with only a few training imagess. It won the \textit{International Symposium on Biomedical Imaging (ISBI)} cell tracking challenge (2015) by training this network on transmitted light microscopy images. Furthermore, this architecture allows fast end-to-end training and inference, and can be trained from only a few training samples~\cite{ronneberger2015u}.
Both architectures, ResNet and U-Net, are state-of-the-art components for image classification and segmentation tasks. 
Anas et al.~\cite{anas2017clinical} presented a residual U-Net (ResUNet), a U-Net that implements residual units, for clinical target-volume delineation in prostate brachytherapy.
In contrast to our work, previous studies only focus on manual or semi-automated analysis of HRF. In the literature we find manual analysis of HRF in SD-OCT scans~\cite{christenbury2013progression}, where HRF was correlated with disease progression of age-related macular degeneration (AMD). 
Korot et al.~\cite{korot2016algorithm} developed a semi-automated pipeline to perform quantification of vitreous HRF in SD-OCT scans. They evaluated the repeatability of the algorithm by comparing the results on successive OCT scans of the same patients. Since Korot et al. did not leverage machine learning methods, no training step was involved. Furthermore, no pixel-level ground-truth data was used for performance analysis of the algorithm.

\paragraph{Contribution}

In this paper, we propose to leverage deep neural networks for fullly automated segmentation of HRF in spectral-domain OCT (SD-OCT) scans. We utilize a ResUNet for accurate HRF segmentation, described in Section~\ref{sec:methods:sem_seg}.
We perform comprehensive evaluation of different semantic segmentation architectures, and the influence of different training objectives (cross-entropy loss vs. Dice loss), and single vendor vs. joint vendor training.
Experiments (Section~\ref{sec:experiments}) on labeled data, extracted from SD-OCT scans, show that this approach segments HRF with high accuracy.
To the best of our knowledge, this is the first published work on fully automated detection and segmentation of HRF.

\section{Semantic segmentation of hyperreflective foci}

Image classification describes the task of classifying the entirety of pixels of an image into a single class out of a prespecified number of object classes, depending on the main object visible in the image. The classification result will most likely conform to the class label of the most prominent image object. In contrast, semantic segmentation~\cite{mostajabi2015feedforward,noh2015learning,long2015fully,chen2015semantic,zheng2015conditional}
performs classification on each pixel of an image in a single pass and thus allows not only to detect but also to localize multiple objects in an image. This can be very run-time efficient, as opposed to performing image segmentation based on an image-level classification approach. The latter is time consuming, because it involves extraction of multiple small image patches, classifying these patches and finally aggregating the classification outputs into a a pixel-level classification map, i.e. the segmentation of the whole image. In the following, we describe the semantic segmentation approach in more detail.

\subsection{Data representation}
\label{sec:methods:representations}
The data comprises $N$ tuples of medical images and pixel-level ground truth annotations
$\langle \mathbf{I}_n, \mathbf{L}_n \rangle$, with $n=1,2,\dots ,N$, where $\mathbf{I}_n \in \mathbb{R}^{a \times b}$ is an intensity image of size $a \times b$ and
$\mathbf{L}_n \in \{0,1\}^{a \times b}$ is a binary image of the same size carrying the information about the pixel-level presence of the object of interest (in our case, a pixel value of 1 indicates the occurrence of HRF in an image on pixel-level).
We extract $K$ small 2D image patches $x_{k,n}$ of size $\dot{a} \times \dot{b}$, with $\dot{a}<a$ and $\dot{b}<b$, from each image $\mathbf{I}_n$ at randomly sampled positions. We extract analogously 2D patches $y_{k,n}$ of the same size from the corresponding positions of the annotation image $\mathbf{L}_n$, resulting in data $\langle \mathbf{x}=x_{k,n}, \mathbf{y}=y_{k,n} \rangle$, with $k=1,2,\dots ,K$.
The overall data is divided into disjoint sets, used for training, validation, or testing of the model.

\subsection{Semantic segmentation methodology}
\label{sec:methods:sem_seg}
\paragraph{Semantic segmentation}
We leverage deep learning based \textit{semantic segmentation} to obtain a mapping from intensity images to corresponding images of dense pixel-level class labels. The underlying feed-forward neural network comprises two main building blocks, which are jointly trained. First, an \textit{encoder} transforms the input image into a low-dimensional abstract context representation. Secondly, a \textit{decoder} maps the low-dimensional embedding, i.e. the output of the encoder, to a full input resolution image of corresponding class label predictions. The most basic processing units of encoder and decoder are convolutional layers. Typically, the encoder produces successively smaller resolution feature maps through the utilization of strided convolutions or convolution with stride 1 followed by a pooling-layer. The decoder produces successively larger resolution images through the utilization of the unpooling operation~\cite{noh2015learning} or implementing fractionally strided convolutions~\cite{radford2015unsupervised,dumoulin2016guide}. The network is trained end-to-end, i.e. parameter updates in every update iteration are based on tuples of intensity images and corresponding images of target labels.

\paragraph{Residual U-Net based semantic segmentation}
The \textbf{\textit{U-Net}} architecture is based on a contracting path (encoder) and an expanding path (decoder). The main contribution of Ronneberger et al.~\cite{ronneberger2015u} in the conception of the U-Net architecture is the concatenation of the feature maps of every layer of the encoder with the feature maps of the corresponding level of the decoder. In this way, higher resolution context information can be propagated to the last decoder layers, which improves localization and allows precise segmentation.
%
The main building blocks of \textbf{\textit{ResNet}} are residual units~\cite{he2016deep}. Residual units not only learn the mapping from inputs to outputs but also learn residual functions between inputs and outputs of individual layers and thus allow even very deep networks learning the identity mapping. Residual units implement ``shortcut connections'', which perform the identity mapping by skipping one or more layers. The outputs of the shortcut connections are added the the outputs of the skipped layers. Since shortcut connections do not add further model parameters, nor increase the computational complexity, end-to-end training of even very deep networks by stochastic gradient descent (SGD) is enabled~\cite{he2016deep}.

Both architectures can be combined to build up a residual U-Net (\textbf{\textit{ResUNet}}), a deep neural network for semantic segmentation with an U-Net architecture with residual units as individual layers.

During training, we learn the mapping $M(\mathbf{x}) = \mathbf{x} \mapsto \mathbf{y}$ by training a deep neural network $M$. During testing, the model $M$ yields images $\mathbf{p}$ of dense pixel-level predictions for unseen testing images $\mathbf{x}_u$.

\subsection{Training objectives}
\label{sec:methods:ojectives}
We train the networks on two objective functions, \textit{cross entropy loss} and \textit{dice loss}.

\subsubsection{Cross entropy loss}
The cross entropy loss is widely used objective function in classification problems. A pixel-level cross-entropy is computed between network predictions $p_{i,j}$ and target labels $y_{i,j}$:
\begin{equation}\label{eqn:ce_loss}
	L_i =  -\sum_j y_{i,j} \log (p_{i,j}),
\end{equation}
where $i$ is the i-th output element (i.e. output at a single pixel) and $j$, with $j=1,2,\dots ,J$ denotes the class. The predictions for this multi-class definition of the cross entropy loss are computed with a pixel-wise softmax function applied on the network outputs $z_{i,j}$:
\begin{equation}\label{eqn:softmax}
	p_{i,j} = \frac{e^{z_{i,j}}}{\sum_{\gamma=1}^J e^{z_{i,\gamma}}}.
\end{equation}
The softmax function maps a vector of arbitrary real valued network outputs $z_{i,j}$ to a vector of values ranging from 0 to 1 that sum to 1.
At each pixel location, the cross entropy loss penalizes the deviation of the network prediction from the ground truth labels.
In the binary case, the cross entropy loss is defined as:
\begin{equation}\label{eqn:binary_ce_loss}
	L_i =  -y \log(p_i) - (1-y_i) \log(1-p_i),
\end{equation}
between targets $y_i$ and class probabilities $p_i$, which are the sigmoidal outputs of a neural network:
\begin{equation}\label{eqn:sigmoid}
	\sigma(z_i) = \frac{1}{1 + e^{-z_i}}
\end{equation}

\subsubsection{Dice loss}
The S{\o}rensen–Dice index~\cite{sorensen1948method,dice1945measures} (another commonly used denotation is \textit{Dice similarity coefficient (DSC)}) is a performance measure used for binary classifier performance evaluation. For binary segmentation problems, the DSC can be utilized to quantify the ``similarity'' of predicted and the true segmentation, and is defined by:
\begin{equation}\label{eqn:dice_score}
	DSC = \frac{2 \cdot t^+}{2 \cdot t^+ + f^+ + f^-},
\end{equation}
where $t^+$ is the number of \textit{true positives}, $f^-$ is the number of \textit{false negatives}, and $f^+$ is the number of \textit{false positives}. Possible values of DSC range from 0.0 to 1.0. A perfect classifier or segmentation model achieves a DSC of 1.0. For binary classification or segmentation problems, the DSC can not only be utilized to evaluate the performance of a trained model on the test set, but also as objective function during training.
When a model is trained to minimize the objective function, a smooth Dice loss can be defined as follows:
\begin{equation}\label{eqn:ce_loss}
	DL = 1.0 - \frac{2 \cdot \sum_i (y_i \cdot p_i) + \epsilon}{\sum_i y_i + \sum_i p_i + \epsilon},
\end{equation}
between real valued network predictions $\mathbf{p}$ and binary target labels $\mathbf{y}$, where $i$ is the i-th pixel.

\section{Experiments}
\label{sec:experiments}

\paragraph{Data, Data Selection and Preprocessing}
We trained and evaluated the method on clinical SD-OCT scans of the retina acquired with devices of two different OCT vendors (\textit{Cirrus} HD-OCT, Carl Zeiss Meditec, and \textit{Spectralis} OCT, Heidelberg Engineering). \textit{Cirrus} scans comprise image slices (\textit{``B-scans``}) with an image resolution of $1024 \times 512$ pixels (pixel sizes $1.96 \times 11.74\mu m$), whereas \textit{Spectralis} comprise 49 \textit{B-scans} with an image resolution of $496 \times 512$ pixels (pixel sizes $3.87 \times 11.23\mu m$) in row-, and column-direction, respectively. Since images of both vendors show different numbers of image rows (i.e. different pixel sizes in row-direction), we rescale the images of both vendors to an uniform image resolution of $320 \times 512$ pixels, resulting in row-direction to a pixel size of $6.26\mu m$ for Cirrus images and of $6.00\mu m$ for Spectralis images. This row-dimension choice was a trade-off between keeping annotation information and moving the pixels beyond isotropy. As a second preprocessing step, the gray values were normalized on a OCT scan basis to range from 0  to 1.
The overall dataset comprises 145 OCT scans from different retinal diseases including age-related macular degeneration (AMD, 60 scans), diabetic macular edema (DME, 43 scans) and retinal vein occlusion (RVO, 42 scans). To keep annotation effort in check, approximately only every $10^{th}$ image of each OCT scan was annotated by clinical retina experts. For training and evaluation, we took only those images that had at least a single pixel annotated as HRF. The dataset was split into a training set (119 OCT scans, 1051 images), a validation set (6 OCT scans, 41 images), and a test set (20 OCT scans, 137 images). Table ~\ref{tab:result/data_stats} lists full details on the data split regarding OCT scans and on to the number of images (\textit{B-scans}). The split was performed on a patient basis so that images of each patient are only assigned to one of these sets. For training we extracted image patches of size $128 \times 32$ pixels at randomly sampled positions.\\

In addition to the OCT scans, used for training, validation, and testing, we had images extracted from 3 additional OCT scans (2 Cirrus and 1 Spectralis) annotated independently by two clinical retina experts. Based on these cases, we evaluate the inter-rater variability as baseline for achievable maximal accuracy.

\begin{table}[tp]
\centering
\caption{Data split statistics. Training set, validation set, and test set with total number of OCT scans ($\#OCT$), total number of images ($\#Img$), number of scans per disease (AMD, DME, RVO), number of Cirrus OCT scans ($\#OCT_{Cir}$), and number of Spectralis OCT scans ($\#OCT_{Spe}$).}
    \begin{tabular}
    { @{}L{.13\textwidth}  C{0.13\textwidth}  C{0.13\textwidth}  C{0.09\textwidth}  C{0.09\textwidth}  C{0.09\textwidth}  C{0.13\textwidth}  C{0.13\textwidth} @{} }
     \hline
 \,Data split     & $\#OCT$ & $\#Img$  & AMD & DME & RVO & $\#OCT_{Cir}$ & $\#OCT_{Spe}$ \\ \hline
 \,Train            &119	         &1051	      & 51 & 33 & 35  &69	                  &50\\
 \,Validation     &6 	         &41	          & 3   &   2 & 1    &3                      &3  \\
 \,Test             &20           &137  	      & 6   &   8 & 6    &10	                  &10 \\ \hline
 \,Total            &145         &1229  	  & 60 & 43 & 42  &82	                  &63 \\
 \hline
    \end{tabular}
\normalsize
\label{tab:result/data_stats}
\end{table}

\subsection{Evaluation}
\label{sec:exp_eval}
HRF appear in OCT data as bright spots, and their segmentation can be formulated as pixel-wise binary classification problem. We examine whether state-of-the art semantic segmentation models (ResUNnet) are required or the application of if even a simple approach suffices. We evaluate the semantic segmentation performance of the following three model architectures:

\begin{enumerate}
\item{\textbf{SemSeg}} is a simple semantic segmentation model with convolutional layers as main units, where the encoder and decoder comprise four layers with $(16-64-64-128)$ and $(128-64-64-16)$ filters of size $(3 \times 3)$ pixels, respectively.
\\
\item{\textbf{ResUnet}} is a residual U-Net with basically the same number of layers (and feature maps per layer) as the \textit{SemSeg} model, i.e., the encoder and decoder comprise four layers with $(16-64-64-128)$ and $(128-64-64-16)$ filters of size $(3 \times 3)$ pixels, respectively. In this model, the standard convolutional layers of the \textit{SemSeg} model are replace with 1 residual unit per scale.
\\
\item{\textbf{ResUnet$^+$}} is similar (same number of layers) to the \textit{ResUnet} model but is a slightly more complex. The encoder and decoder comprise four layers with $(32-64-128-256)$ and $(256-128-64-32)$ filters of size $(3 \times 3)$ pixels, respectively. In the \textit{ResUnet$^+$} model, we use 3 residual units per layer.
\end{enumerate}

To examine the influence of training objectives on the segmentation performance, each of these models is trained on a \textbf{(1)} \textit{cross entropy loss} or \textbf{(2)} \textit{Dice loss}.
Receiver operating characteristic (ROC) curves can present a possibly misleading optimistic visualization of the model performance if the class distribution of the data has a strong skew~\cite{davis2006relationship}. Precision-recall curves are an alternative to ROC curves on data with high class imbalance~\cite{goadrich2004learning,bunescu2005comparative,craven2005markov}. Since our data comprises highly imbalanced classes, i.e. the coverage of HRF even in positive OCT scans is relative small, we use \textit{precision-recall curves} as summary statistic to visualize the model performance. We report the \textit{average precision (AP)} to quantitatively summarize the precision-recall curve. We report area under the ROC curve (AUC) values for the sake of completeness only.

\paragraph{Implementation details}
For training the ResUNet models, we utilized the \textit{Deep Learning Toolkit for Medical Imaging (DLTK)}~\cite{pawlowski2017dltk}.
All models were trained for 200 epochs, and model parameters were stored at the best performing epoch on the validation set. After model selection and hyperparameter tuning, the final performance was evaluated on the test set using the learned model parameters. We utilized the stochastic optimizer Adam~\cite{kingma2014adam} during training. All experiments were performed using Python 2.7 with the TensorFlow~\cite{tensorflow2015-whitepaper} library version 1.2, CUDA 8.0, and a Titan X graphics processing unit.

\subsection{Results}
\begin{figure}[htp]
  \centering
     \includegraphics[width=1\textwidth]{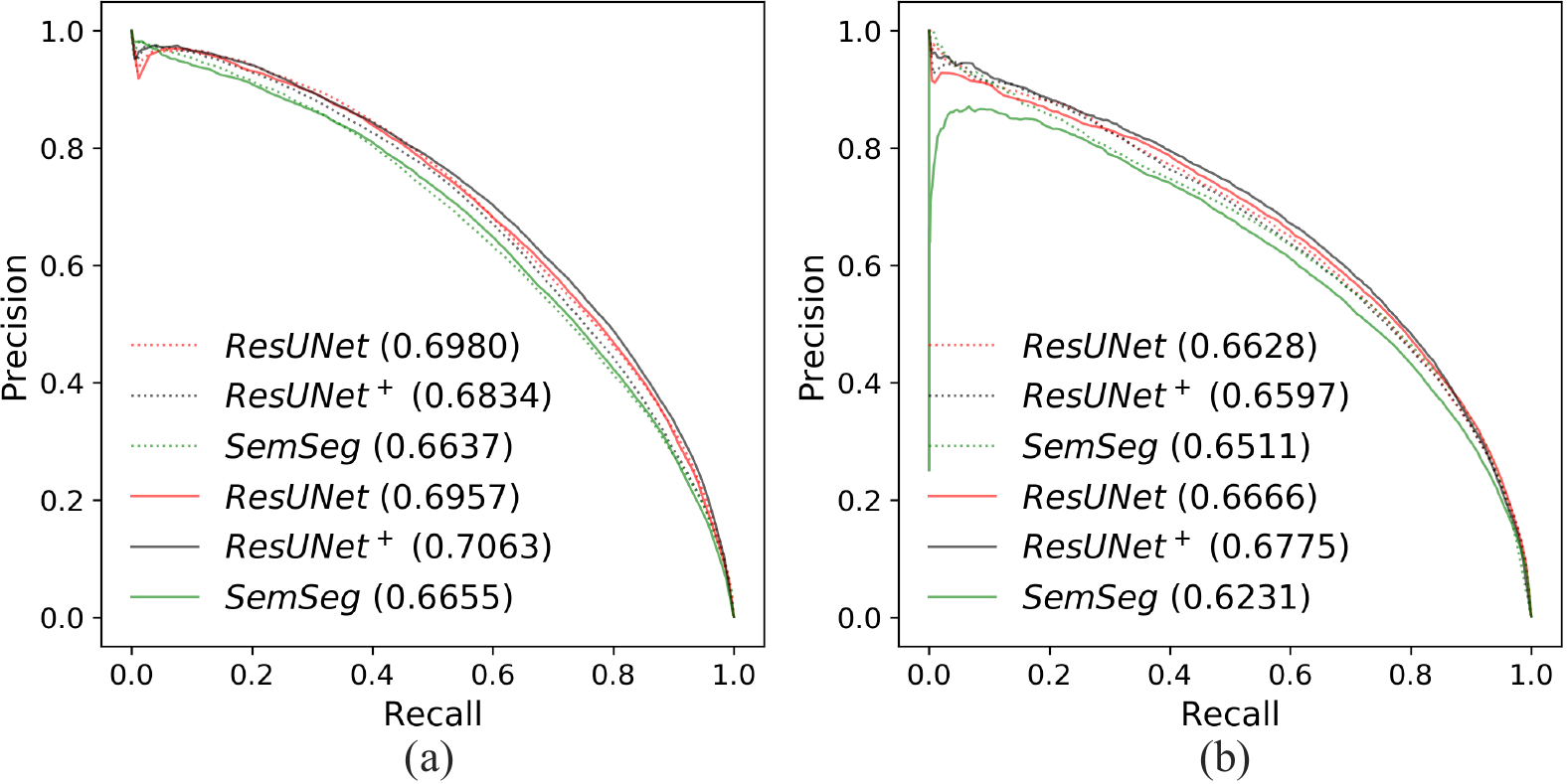}
     \vspace{-5mm}
  \caption{HRF segmentation performance evaluation. a) Precision-recall curves for different models tested on \textit{Cirrus data}. Training on Cirrus data only (solid line) or jointly training on Cirrus and Spectralis. b) Precision-recall curves for different models tested on \textit{Spectralis data}. Training on Spectralis data only (solid line) or jointly training on Cirrus and Spectralis. Corresponding AP values are given in parenthesis.}
  \label{fig:result/PRc}
\end{figure}

\begin{figure}[htp]
  \centering
     \includegraphics[width=1\textwidth]{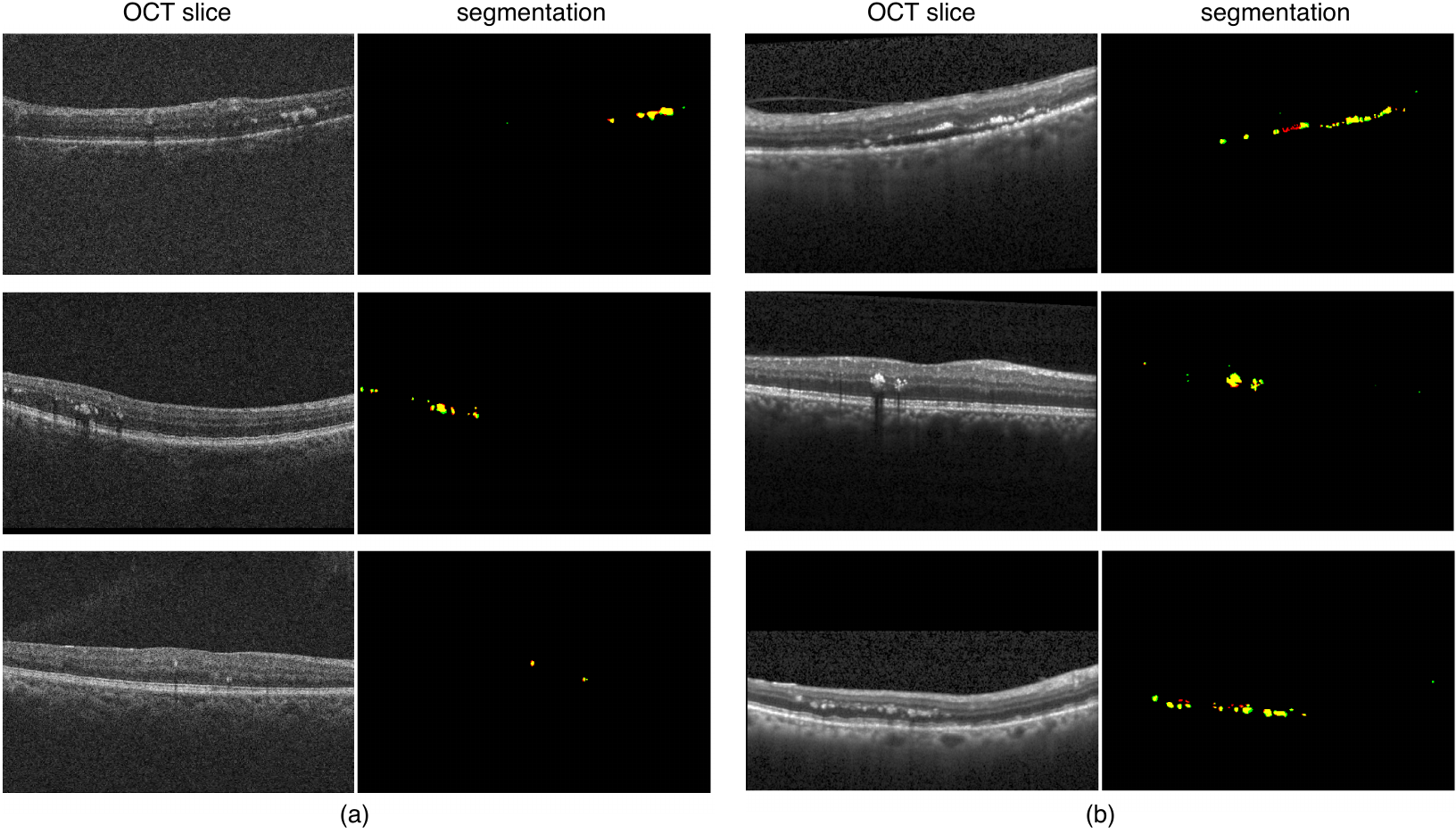}
  \caption{HRF segmentation. (a) Cirrus OCT data. (b) Spectralis OCT data. Bscans (left) and corresponding segmentation results (right). True positives (yellow), false positives (green), false negatives (red), and true negatives (black).}
\label{fig:result/Segmentation}
\end{figure}

Results demonstrate the applicability of all examined semantic segmentation algorithms.
Detailed quantitative results are listed in \Crefrange{tab:result/cirrus_single}{tab:result/multivend_ce}, which show the ResUNet$^+$ model jointly trained on Cirrus and Spectralis data utilizing a cross-entropy loss is the best performing model. This observation holds true for testing data of both vendors, with the best AP on Cirrus data of $0.7063$ (DSC of $0.6526$) and with the best AP on Spectralis data of $0.6775$ (DSC of $0.6349$).
The superiority of the ResUNet$^+$ model is also evident through the precision-recall curves shown in Figure~\ref{fig:result/PRc}. Qualitative segmentation results of this model are shown for Cirrus OCT data in~\Cref{fig:result/Segmentation}(a) and for Spectralis OCT data in~\Cref{fig:result/Segmentation}(b).

\paragraph{Comparison to the inter-rater variability}
In addition to the OCT scans, used for training, testing, and evaluation, we had 25 images extracted from 3 additional OCT scans (2 Cirrus and 1 Spectralis) annotated independently by two clinical retina experts. The DSC over the double annotated images is $0.6760$, which is a measure for the \textit{inter-rater variability}.
The average accuracy (DSC) of the best performing model (ResUNet$^+$) over all test images of Cirrus scans and Spectralis scans is $0.6430$ and thus lies in the order of magnitude of the inter-rater variability. Even-though we only had a limited number of double annotated images, results suggest that the presented approach is highly accurate.

\begin{table}[H]
\centering
\caption{Training and testing on Cirrus data utilizing a cross entropy loss (\textit{CE}) or utilizing a Dice-based loss (\textit{Dice}). Note that we report AUC for completeness, while AP is a more relevant measure in this case due to the strong skewness of the label distribution.}
\vspace{-2mm}
    \begin{tabular}
    { @{}L{.1\textwidth} L{.15\textwidth}  C{0.13\textwidth}  C{0.13\textwidth}  C{0.13\textwidth}  C{0.13\textwidth} C{0.13\textwidth} }
Loss & Model & Precision & Recall & DSC & AP &AUC \\ \hline
~                 & SemSeg   & 0.585  & 0.6502 & 0.6159 & 0.6637 &0.9978 \\
\textit{CE}   & ResUNet  & 0.7126 & 0.5704 & 0.6336 & 0.6980 &0.9956 \\
~                 & ResUNet+ & 0.6692 & 0.6021 & 0.6339 & 0.6834 &0.9939\\ \hline \hline
~                 & SemSeg   & 0.5932 & 0.6236 & 0.6080  & 0.5489 &0.9515 \\
\textit{Dice} & ResUNet  & 0.5513 & 0.7149 & 0.6226 & 0.5431 &0.9556 \\
~                 & ResUNet+ & 0.5832 & 0.6416 & 0.6110  & 0.5318 &0.9467
    \end{tabular}
\normalsize
\label{tab:result/cirrus_single}
\vspace{-6mm}
\end{table}

\begin{table}[H]
\centering
\caption{Training and testing on Spectralis data utilizing a cross entropy loss (\textit{CE}) or utilizing a Dice-based loss (\textit{Dice}).}
\vspace{-2mm}
    \begin{tabular}
    { @{}L{.1\textwidth} L{.15\textwidth}  C{0.13\textwidth}  C{0.13\textwidth}  C{0.13\textwidth}  C{0.13\textwidth} C{0.13\textwidth} }
Loss & Model & Precision & Recall & DSC & AP &AUC \\ \hline
~                 & SemSeg   & 0.5099 & 0.7585 & 0.6099 & 0.6511 &0.9869 \\
\textit{CE}   & ResUNet  & 0.5318 & 0.7348 & 0.6171 & 0.6628 &0.9961 \\
~                 & ResUNet+ & 0.547  & 0.7137 & 0.6193 & 0.6597 &0.9961\\ \hline \hline
~                 & SemSeg   & 0.5269 & 0.7042 & 0.6028 & 0.5663 &0.9269 \\
\textit{Dice} & ResUNet  & 0.4963 & 0.7334  & 0.592   & 0.5429 &0.9222 \\
~                 & ResUNet+ & 0.6245 & 0.6742 & 0.6484 & 0.5854 &0.9325
    \end{tabular}
\normalsize
\label{tab:result/spe_single}
\vspace{-6mm}
\end{table}

\begin{table}[H]
\centering
\caption{Jointly trained on Cirrus and Spectralis data utilizing a cross entropy loss (\textit{CE}). Testing on \textit{Cirrus} data or testing on \textit{Spectralis} data.}
\vspace{-2mm}
    \begin{tabular}
    { @{}L{.15\textwidth} L{.15\textwidth}  C{0.13\textwidth}  C{0.13\textwidth}  C{0.13\textwidth}  C{0.13\textwidth} C{0.11\textwidth} }
Test data & Model & Precision & Recall & DSC & AP &AUC \\ \hline
~                     & SemSeg   & 0.6455 & 0.6031 & 0.6236 & 0.6655 &0.9928 \\
\textit{Cirrus}  & ResUNet  & 0.6256 & 0.6624 & 0.6434 & 0.6957 &0.9954 \\
~                     & ResUNet+ & 0.6655 & 0.6401 & \textbf{0.6526} & \textbf{0.7063} &0.9915\\ \hline \hline
~                     & SemSeg   & 0.4668 & 0.7688 & 0.5809 & 0.6231 &0.9962 \\
\textit{Spectralis} & ResUNet  & 0.5309 & 0.7497 & 0.6216 & 0.6666 &0.9951 \\
~                     & ResUNet+ & 0.5598 & 0.7332 & \textbf{0.6349} & \textbf{0.6775} &0.9917
    \end{tabular}
\normalsize
\label{tab:result/multivend_ce}
\vspace{-6mm}
\end{table}

\section{Conclusion} 
We applied semantic segmentation for fully automated segmentation of HRF in retinal OCT scans. This is the first time that a fully automated method for the detection and segmentation of HRF is reported. Based on results of all experiments, evidence suggests, that the utilization of cross entropy training loss should be given preference over a Dice-based training objective. Results demonstrate the general applicability of all examined semantic segmentation algorithms. However, ResU-Nets are able to detect HRF with slightly higher accuracy and handle the visual variability of input images best, i.e. joint training on OCT scans of different vendors. Since we trained one model on data acquired with different devices (i.e., vendors) from patients with different retinal diseases including AMD, DME and RVO, -- as against training specific models on individual diseases -- the algorithm can safely be applied for screening in all of them even if the pathophysiological origins are different.

\section*{Acknowledgements}

This work has received funding from IBM, FWF (I2714-B31), OeNB (15356, 15929), the Austrian Federal Ministry of Science, Research and Economy (CDL OPTIMA). We gratefully acknowledge the support of NVIDIA Corporation with the donation of a GPU used for this research.

\bibliographystyle{splncs}
\bibliography{references}
\end{document}